\documentclass[10pt,journal,compsoc]{IEEEtran}
\usepackage{amsmath,amsfonts}
\usepackage{algorithmic}
\usepackage{array}
\usepackage[caption=false,font=normalsize,labelfont=sf,textfont=sf]{subfig}
\usepackage{textcomp}
\usepackage{stfloats}
\usepackage{url}
\usepackage{verbatim}
\usepackage{graphicx}
\usepackage{cite}
\usepackage[linesnumbered,ruled,lined]{algorithm2e}
\usepackage{amsthm}
\usepackage{epstopdf}
\usepackage{multirow}
\usepackage{booktabs}
\usepackage{diagbox}

\newtheorem {lemma} {\bf Lemma}

\newtheorem {theorem} {\bf Theorem}[]

\begin{document}
%

\title{C$^{2}$IMUFS: Complementary and Consensus Learning-based Incomplete Multi-view Unsupervised Feature Selection}
%
\author{Yanyong~Huang,~Zongxin~Shen,~Yuxin~Cai,~Xiuwen Yi,~Dongjie~Wang,~Fengmao~Lv~and~Tianrui~Li,~\IEEEmembership{Senior Member,~IEEE}
\IEEEcompsocitemizethanks{\IEEEcompsocthanksitem Yanyong~Huang, Zongxin~Shen and Yuxin~Cai are with the School of  Statistics, Southwestern University of  Finance and Economics, Chengdu 611130, China (e-mail: huangyy@swufe.edu.cn; shenzx@smail.swufe.edu.cn; caaaiyx@163.com);
\IEEEcompsocthanksitem Xiuwen~Yi is with the JD Intelligent Cities Research and JD Intelligent Cities Business Unit, Beijing 100176, China (e-mail: xiuwenyi@foxmail.com);
\IEEEcompsocthanksitem Dongjie~Wang is with the College of Engineering and Computer Science, University of Central Florida, Orlando 32816, Florida (e-mail: wangdongjie@knights.ucf.edu);
\IEEEcompsocthanksitem Fengmao~Lv and Tianrui~Li   are with the School of Computing and Artificial Intelligence, Southwest Jiaotong University, Chengdu 611756, China (e-mail: fengmaolv@126.com; trli@swjtu.edu.cn).}
}


\markboth{Journal of \LaTeX\ Class Files,~Vol.~14, No.~8, August~2021}%
{Shell \MakeLowercase{\textit{et al.}}: Bare Demo of IEEEtran.cls for Computer Society Journals}

\IEEEtitleabstractindextext{%
\begin{abstract}
Multi-view unsupervised feature selection (MUFS) has been demonstrated as an effective technique to reduce the dimensionality of multi-view unlabeled data. The existing methods assume that all of views are complete. However, multi-view data are usually incomplete, i.e., a part of instances are presented on some views but not all views. Besides, learning the complete similarity graph, as an important promising technology in existing MUFS methods, cannot achieve due to the missing views. In this paper, we propose a complementary and consensus learning-based incomplete multi-view unsupervised feature selection  method (C$^{2}$IMUFS) to address the aforementioned issues. Concretely, C$^{2}$IMUFS  integrates feature selection into  an extended weighted non-negative matrix factorization model equipped with adaptive learning of view-weights and a sparse $\ell_{2,p}$-norm, which can offer better adaptability and flexibility. By the sparse linear combinations of multiple similarity matrices derived from different views, a complementary learning-guided similarity matrix reconstruction model is presented to obtain the complete similarity graph in each view. Furthermore,  C$^{2}$IMUFS learns a consensus clustering indicator matrix across different views and embeds it into a spectral graph term to preserve the local  geometric structure. Comprehensive experimental results on real-world datasets demonstrate the effectiveness of C$^{2}$IMUFS compared with state-of-the-art methods. 
\end{abstract}

\begin{IEEEkeywords}
Unsupervised feature selection, incomplete multi-view data, complementary and consensus information, weighted non-negative matrix factorization.
\end{IEEEkeywords}}

\maketitle

\IEEEdisplaynontitleabstractindextext

%
\IEEEpeerreviewmaketitle

\IEEEraisesectionheading{\section{Introduction}\label{sec:introduction}}
\IEEEPARstart{M}{ulti-view} data widely exist in many real applications, where the same instances are described by multiple heterogeneous feature sets from different aspects. For example, in the video classification task, each video can be depicted by the visual features, audio inputs and text representations~\cite{Z.X.Wu2014}. In the credit risk assessment task, Bad Credit History, Loan Description and Extra Credit Information are often used to describe each loan application from different perspectives~\cite{CuiL2021}. In these applications, different views of data are usually represented in high dimensional feature space containing redundant and noisy features, which will result in the problem of ``curse of dimensionality'' and degrade the performance of clustering or classiﬁcation tasks~\cite{Z.Y.Guan2015}. In addition, it is difficult or even infeasible to obtain a large amount of labeled data in practice. Therefore, how to reduce the dimensionality of multi-view unlabeled data for the improvement of performance of the subsequent tasks has become an urgent problem to be solved  in various practical applications.

Multi-view unsupervised feature selection (MUFS) provides an effective solution to address the above problem by choosing a compact subset of representative features from the original feature space. In recent years, a variety of MUFS approaches have been developed, which can be mainly classified into two categories. In the first category are the traditional single-view unsupervised feature selection methods, which first concatenate all features derived from different views and then take the combined features as input straightforward~\cite{LapScore,RSFS,UDFS,CNAFS}. There are some typical single-view based unsupervised feature selection methods, like Laplacian Score (LPscore)~\cite{LapScore}, Robust Spectral Feature Selection (RSFS)~\cite{RSFS} and  Embedded Graph learning and Constraint-based Feature Selection (EGCFS)~\cite{EGCFS}. These methods usually characterize the local geometric structure of data by constructing various graphs and select the top ranked features according to their importance. Although these methods can help perform better feature selection to a certain extent, they treat each view independently and do not consider the underlying correlations among different views. Rather than combining features of different views, the second category of MUFS methods directly construct models from multi-view data  for feature selection~\cite{AMFS,ASVW,MUFSurvey}. Typical methods in this category include  Adaptive Collaborative Similarity Learning (ACSL)~\cite{ACSL}. ACSL adaptively learns a collaborative similarity structure of data by utilization of complementary information across different views and integrates it into a sparse regression model for feature selection. Besides, Tan et al. have proposed the method referred to as Cross-view Local structure Preserved Diversity and Consensus Learning (CvLP-DCL), which 
employs a cross-view similarity graph learning to preserve the local geometric structure of data and learns the consensus and diversity label spaces to select features~\cite{CvLP-DCL}. Furthermore, Bai et al. have developed a unified learning framework for multi-view unsupervised feature selection, named Nonnegative Structured Graph Learning (NSGL)~\cite{NSGL}. It learns a common adaptive structured graph among different views and embeds it into the sparse linear regression model to learn the consensus pseudo labels and perform feature selection simultaneously. Although these proposed methods have been demonstrated promising performance in feature selection, they have a common assumption, i.e., each instance appears in all of views.  In real-world applications, it is more often the case that a part of instances are presented on some views but not all views~\cite{J.Wen2021, L.Li2021}. For example, in web image retrieval, the visual information, links, and textual tags of each image can be regarded as three views, while some images may lack the link or text view~\cite{K.H.Guo2016}. Hence, the above-mentioned methods could not be used in the incomplete multi-view data directly. Besides, an important promising technology of MUFS is that learning or constructing the complete similarity graph in each view by employing the complementary information across different views. However, the existing methods can not obtain the complete graphs since some views are missing for part of instances in the incomplete multi-view scenario. Furthermore, these methods do not consider simultaneously using the consensus and complementary information among different views  and treat all views equally ignoring the difference of different views in fusion of multi-view information. As a result,  these will limit their performance in incomplete multi-view unsupervised feature selection.

In order to address the aforementioned issues, we present a novel MUFS method for multi-view unlabeled data with incomplete views, named as Complementary and Consensus Learning-based Incomplete Multi-view Unsupervised Feature Selection (C$^{2}$IMUFS). Specifically, our proposed method C$^{2}$IMUFS embeds the process of feature selection into an extended weighted non-negative matrix factorization model (WNMF), which can automatically determine the importance of different views and consider the difference between the missing and non-missing instances. To obtain the complete similarity graph in the incomplete-view scenario, we propose a complementary learning-guided similarity matrix reconstruction model. It can adaptively learn the similarity matrix by sparse linear combinations of multiple similarity matrices derived from other views. Meanwhile, C$^{2}$IMUFS learns a consensus clustering indicator matrix among multiple views and integrates it into a spectral graph term to preserve the common  geometric structure of different views. Furthermore, a sparse constraint with $\ell_{2,p}$-norm is imposed on the feature selection matrix, which can offer better flexibility and satisfy different sparsity requirements. Finally, an alternative iterative optimization algorithm is developed to solve the proposed model and comprehensive experiments are carried out to demonstrate the effectiveness of the proposed method by comparing with several state-of-the-art single-view and multi-view unsupervised feature selection methods. 

To sum up, the major contributions of this paper are summarized as follows: (\romannumeral1) To best of our knowledge, this is the first study to address the problem of unsupervised feature selection on incomplete multi-view data by simultaneously employing the complementary and consensus information across different views. (\romannumeral2) We propose a novel model that incorporates an extended WNMF module and a complementary and consensus learning-based module for incomplete multi-view unsupervised feature selection. It can learn the complete similarity graph of each view by using the complementary graph structures acquired from other views and integrate different views together via learning a consensus indicator matrix with adaptive view weights. (\romannumeral3) We develop an efficient alternative optimization algorithm to solve the proposed model and 
extensive experimental results on eight real-world datasets show the superiority of the proposed method compared with the state-of-the-art methods.

The remainder of this paper is organized as follows. In Section~\ref{sec:Related work}, we briefly review related work about MUFS. In Section~\ref{sec:proposed method}, we introduce in detail the proposed method C$^{2}$IMUFS. Section~\ref{sec:optimization} provides an effective solution to this method and the convergence and complexity of the proposed algorithm are discussed in the following Section~\ref{sec:Discussions}. In Section~\ref{sec:Experimental}, a series of  experiments are carried out to demonstrate the effectiveness of the proposed method. And finally, the conclusions are presented in Section~\ref{sec:Conclusions}.

\begin{figure*}[!htpb]
	\centering
	\includegraphics[width=\textwidth]{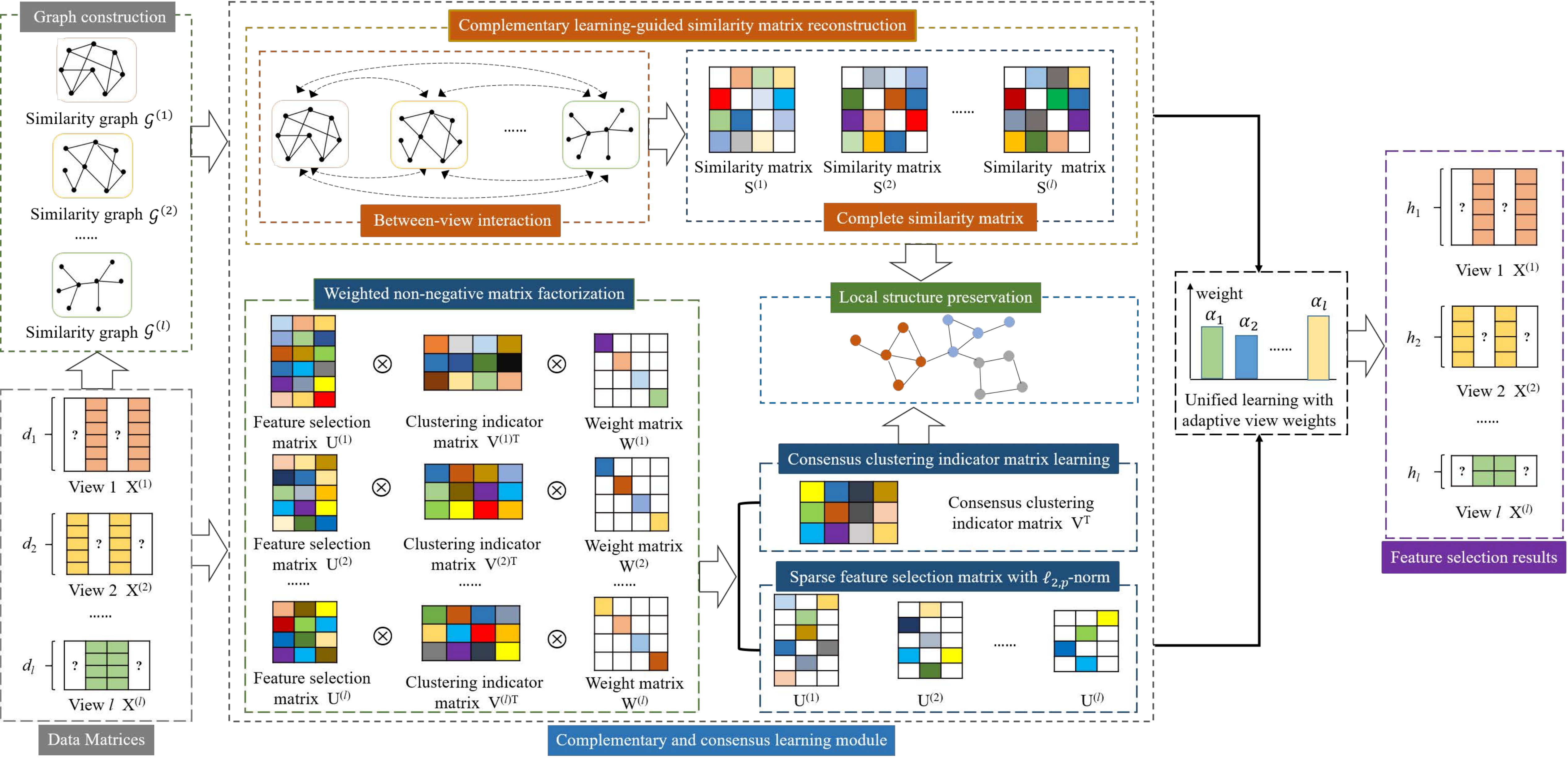}
	\caption{The framework of the proposed Complementary and Consensus learning-based Incomplete Multi-view Unsupervised Feature Selection method (C$^{2}$IMUFS). ? in data matrices denote the missing instances.}\label{Framework}
\end{figure*}

\section{Related Work}\label{sec:Related work}
In this section, we briefly review some representative works in terms of single-view-based unsupervised feature selection and multi-view unsupervised feature selection methods, respectively. 

\subsection{Single-view  Unsupervised Feature Selection Methods}
Single-view  unsupervised feature selection methods are roughly categorized into three classes, i.e., filter-based, wrapper-based and embedded-based  methods. The first kind of methods select the representative features according to certain evaluation criterion  measuring the importance of features. LapScore~\cite{LapScore} selects features in terms of the metric Laplacian Score, which describes the capability of features in preserving  locality.  Zhao et al. employed the spectral graph theory to identify the redundant features~\cite{Z.Zhao2007}. Filter-based methods ignore the relation of selected features and are independent of any learning algorithms, which will result in that they could not obtain the satisfied performance. The second kind of methods usually use a learning algorithm to evaluate the selected features until obtaining the most suitable features. Maldonado et al. developed a wrapper-based feature selection method by employing support vector machines with kernel functions~\cite{S.Maldonado2009}. The main drawback of wrapper methods is their expensive computational cost arising by trying to find the features with the highest performance. The last kind of methods embed the procedure of feature selection into the learning phase of a specific algorithm, which can obtain the important features with low computational cost. RSFS~\cite{RSFS} constructs a graph Laplacian to capture the local geometric structure of data and incorporates the feature selection procedure  into a robust spectral regression model. Yang et al. proposed a unified framework for unsupervised feature selection by combining  the discriminant analysis and $\ell_{2}$-norm minimization~\cite{UDFS}. EGCFS~\cite{EGCFS} embeds the construction of the similarity graph into the procedure of maximizing the between-class scatter matrix, which can adaptively learn the graph structure  and select discriminative features simultaneously. While using these single-view unsupervised feature selection methods to select features from multi-view data,  they directly concatenate all features of different views together and ignore the latent correlations among multiple views. This will result in the inferior performance. 

\subsection{Multi-view  Unsupervised Feature Selection Methods}
Different from the above-mentioned methods, multi-view  unsupervised feature selection methods deal with multi-view data directly and explore the underlying information among different views to improve the performance.  Shao et al. presented  an online unsupervised multi-view feature selection method (OMVFS), which employs the  non-negative matrix factorization model to simultaneously learn the feature selection matrices with sparse constraint and a consensus cluster indicator matrix across different views~\cite{OMVFS}. ACSL~\cite{ACSL} integrates the adaptive collaborative graph learning and feature selection into a unified framework by employing the complementary information among different views and a sparse regression model. CvLP-DCL~\cite{CvLP-DCL} learns a cross-view similarity graph equipped with a matrix-induced regularization term by linearly combining different view-specific similarity graphs. Besides, in CvLP-DCL, a common and view-specific label spaces are incorporated into a regression model for feature selection.  Tang et al. embedded feature selection into a clustering process based on non-negative matrix factorization, which can both obtain a common clustering indicator matrix and these latent feature matrices of different views~\cite{CGMV-UFS}. In NSGL~\cite{NSGL}, the common similarity graph learning among multiple views with adaptive view weights and pseudo label learning are combined into a linear regression model with $\ell_{2,1}$-norm regularization for feature selection. However, the aforementioned  methods are confronted with three issues. First, they all assume that all of views are complete, which disable them to be used in the feature selection of incomplete multi-view data directly. Second, these methods can not obtain the complete similarity graph since the problem of missing views. This will result in the inferior performance. Last, they did not take the complementary and consensus information among different views into consideration simultaneously. These will limit their performance.

\section{Complementary and Consensus Learning-based Incomplete Multi-view Unsupervised Feature Selection}\label{sec:proposed method}
In this section, we first summarize some notations and definitions used throughout the paper. Then, we introduce the proposed method  C$^{2}$IMUFS for incomplete  multi-view  data, which is consist of two parts. The first part employs an extended WNMF model equipped with $\ell_{2,p}$-norm regularization term and adaptive view-weight learning to obtain the feature selection matrices and clustering indicator matrices. In the second part, complementary  and  consistency learning strategies are integrated into the extended model to reconstruct the complete similarity matrix of each view and learn the common clustering indicator matrix. The framework of C$^{2}$IMUFS is shown in Fig. \ref{Framework}.

\subsection{Notations}
In this paper, matrices and vectors are written in boldface uppercase letters and boldface lowercase letters, respectively. Given a matrix $\mathbf{W} \in \mathbb{R}^{m \times n}$, $\mathbf{w}^{j}$ and $\mathbf{w}_{i}$ respectively indicate the  $j$-th row and $i$-th column of matrix $\mathbf{W}$. The $(i, j)$-th entry of $\mathbf{W}$ is denoted by $\mathbf{W}_{ij}$. Let $\operatorname{Tr}(\mathbf{W})$, $\mathbf{W}^{T}$ and $\operatorname{Rank}(\mathbf{W})$ respectively denote the trace of $\mathbf{W}$, the transpose of $\mathbf{W}$ and the rank of $\mathbf{W}$. The Frobenius norm of matrix  $\mathbf{W}$ is defined as $\|\mathbf{W}\|_{F}=\sqrt{\sum_{i=1}^{m}\sum_{j=1}^{n}\left|\mathbf{W}_{i j}\right|^{2}}$.  The $\ell_{2,p}$-norm is defined as $\|\mathbf{W}\|_{2,p}=(\sum_{i=1}^{m}\left\|\mathbf{w}^{i}\right\|_{2}^{p})^{\frac{1}{p}},0<p \leq 1$, where $\left\|\mathbf{w}^{i}\right\|_{2}$ denotes the $\ell_{2}$-norm of vector $\mathbf{w}^{i}$. Especially, while $p=1$, we have the $\ell_{2,1}$-norm: $\|\mathbf{W}\|_{2,1}=\sum_{i=1}^{m} \sqrt{\sum_{j=1}^{n} \mathbf{W}_{i j}^{2}}=\sum_{i=1}^{m}\left\|\mathbf{w}^{i}\right\|_{2}$. We use $\mathbf{I}$ to denote an identity matrix and use $\mathbf{1}$ to indicate a column vector with all elements being one. 

Let $\mathcal{X}=\left\{\mathbf{X}^{(v)},v=1,\ldots,l\right\}$ denote a multi-view data with $l$ views,  where $\mathbf{X}^{(v)} \in \mathbb{R}_{+}^{d_{v} \times N}$ is the non-negative data matrix in view $v$ with $N$ instances and $d_{v}$ features. In real applications, some data instances may be missing in certain views.  To describe  the view-missing information, an indicator matrix $\mathbf{A}=(\mathbf{A}_{ij})_{N \times l}$ is defined as follows:
\begin{equation}\label{1}
\mathbf{A}_{ij}=\left\{\begin{array}{ll}
1 & \text{if the } i\text{-th instance is in the } j \text {-th view;} \\
0 & \text{otherwise.}
\end{array}\right.
\end{equation}
Then, it is easy to know $\sum_{i=1}^{N} \mathbf{A}_{ij} < N$. Our goal is to select $h$ most discriminative features from the incomplete multi-view data $\mathcal{X}$.

\subsection{WNMF-based Incomplete Multi-view Unsupervised Feature Selection}
The traditional WNMF model~\cite{WNMF}  has been widely used to deal with incomplete data, which factorizes a data matrix into two non-negative matrices and introduces a weight matrix to describe the missing information. It can be formulated as follows:
\begin{equation}\label{EqWNMF}
\min _{\mathbf{U}, \mathbf{V}}\|(\mathbf{X}-\mathbf{UV}^{T}) \mathbf{W}\|_{F}^{2},
\end{equation}
where $\textbf{X}$ is an incomplete data matrix,  $\mathbf{W}$ is a weight matrix, and the weighted matrix  entry $\mathbf{W}_{ij}=0$, if the entry $\textbf{X}_{ij}$ of matrix $\textbf{X}$ is missing, otherwise $\mathbf{W}_{ij}=1$.  To select features from the incomplete multi-view data, a direct method is that using Eq.~(\ref{EqWNMF}) to approximately factorize each missing  view data matrix. However, this way is equivalent to fill the missing values with 0 in different views. A more reasonable method is using the weighted averages to fill in the missing values and the weight is determined by  the available instances in each view. To this end, a  WNMF-based incomplete multi-view unsupervised feature selection is proposed as follows:
\begin{equation}\label{NWNMF}
	\begin{aligned}
		&\min _{\{\mathbf{U}^{(v)},\mathbf{V}^{(v)}\}} \sum_{v=1}^{l}\|(\mathbf{X}^{(v)}-\mathbf{U}^{(v)} \mathbf{V}^{(v) T})\mathbf{W}^{(v)}\|_{F}^{2}+\lambda\|\mathbf{U}^{(v)}\|_{\rho} \\
		&{s.t.} \mathbf{V}^{(v)} \geq 0,\mathbf{V}^{(v) T}\mathbf{V}^{(v)}=\mathbf{I},\mathbf{U}^{(v)} \geq 0, v=1,2, \ldots, l,
	\end{aligned}
\end{equation}
where $\mathbf{X}^{(v)}$ is the $v$-th view data matrix, $\mathbf{U}^{(v)}\in \mathbb{R}_{+}^{d_{v} \times c}$ and $\mathbf{V}^{(v)}\in \mathbb{R}_{+}^{N \times c}$  respectively denote the $v$-th view feature selection matrix and clustering indicator matrix,  $\|\mathbf{U}^{(v)}\|_{\rho}$ is the regularization notation of $\mathbf{U}^{(v)}$, $\lambda$ is the regularization parameter, and $c$ is the number of clusters. In Eq.~(\ref{NWNMF}), $\mathbf{W}^{(v)}\in \mathbb{R}_{+}^{N \times N}$  is a diagonal weight matrix in the $v$-th view, whose diagonal entry $\mathbf{W}_{jj}^{(v)}$  is defined as follows:
\begin{equation}\label{3}
	\mathbf{W}_{jj}^{(v)}=\left\{\begin{array}{l}
		1 \quad \text { if instance } j \text { exists in view } v ; \\
		\frac{\sum_{i=1}^{N} \mathbf{A}_{iv}}{N} \quad \text {otherwise. }
	\end{array}\right.
\end{equation}
Eq.~(\ref{3}) shows that the weights of non-missing instance and  missing instance are 1 and the ratio of all available instances in the $v$-th view, respectively. Hence, it can better measure the quality of available information in each view and give a more reasonable weighted average to impute the missing value.

To achieve feature selection, $\ell_{2,1}$-norm is usually used to impose row sparsity  on the $v$-th feature selection matrix $\mathbf{U}^{(v)}$ in Eq.~(\ref{NWNMF})~\cite{OMVFS,M.J.Qian2014}. Based on $\ell_{2,1}$-norm, we can select the top ranked features according to the descending order of  $\ell_{2}$-norm of the rows in $\mathbf{U}^{(v)}$. However, many studies have demonstrated that $\ell_{2,p}$-norm ($0<p<1$) can give better sparsity in comparison with $\ell_{2,1}$-norm~\cite{R.Chartrand2007,L.P.Wang2014,H.Tao2015}. Hence, in Eq.~(\ref{NWNMF}), we impose $\ell_{2,p}$-norm regularization on  $\mathbf{U}^{(v)}$ to ensure its row-sparsity. Then, we can provide more flexible $p$ to satisfy the different sparsity requirements.  Furthermore, considering that different views have different contributions to the final feature selection task,  we use an adaptive weight determination strategy to assign the view-weight  $\alpha^{(v)} (v=1,2,\cdots,l)$ for each view. Then, we can obtain the following novel incomplete multi-view unsupervised feature selection method based on WNMF.
\begin{equation}\label{NWNMF2}
	\begin{aligned}
		&\min _{\{\mathbf{U}^{(v)}\}, \{\mathbf{V}^{(v)}\}} \sum_{v=1}^{l}(\alpha^{(v)})^{\gamma}[\|(\mathbf{X}^{(v)}-\mathbf{U}^{(v)} \mathbf{V}^{(v) T})\mathbf{W}^{(v)}\|_{F}^{2}\\
		& +\lambda\|\mathbf{U}^{(v)}\|_{2,p}^{p}]\\
		&{s.t.~} \mathbf{V}^{(v)} \geq 0,\mathbf{V}^{(v) T}\mathbf{V}^{(v)}=\mathbf{I},\mathbf{U}^{(v)} \geq 0,  0 \leq \alpha^{(v)} \leq 1,\\ 
		& \sum_{v=1}^{l} \alpha^{(v)}=1, v=1,2, \ldots, l,
	\end{aligned}
\end{equation}
where $\gamma$ is a regularization parameter. 

In Eq.~(\ref{NWNMF2}), the proposed model can select features from the incomplete multi-view unlabeled data. Nevertheless, it is limited by the inability to fully explore the inherent information of multi-view data.  In what follows, we will exploit the complementary and consistent information of multi-view data and integrate them into our model to improve its performance. 

\subsection{Complementary and Consensus Learning}
Previous studies have confirmed that preserving local manifold structure by constructing a nearest neighbor graph is beneficial to improve the performance of unsupervised feature selection~\cite{H.J.Wan2019,Y.Y.Huang2021,Z.C.Li2013}. Hence, for the construction of  graph in incomplete multi-view data, the key point is to construct the  similarity matrix in each view. However, it is hard to obtain the reliable similarity matrix in each view since some instances are missing in certain views. Considering that there exists complementary information across different views~\cite{H.Yu2020,C.Q.Zhang2018} and the similarity relation between two instances exists in at least one view, we can construct the similarity matrix by using the between-view interaction similarity information.  Based on this, we propose the following complementary learning-guided similarity matrix reconstruction model, which can adaptively learn the similarity matrix by means of the similarity information derived from other views.  
\begin{equation}\label{5}
\begin{aligned}
&\min _{\{\mathbf{S}^{(v)}\}, \mathbf{R}} \sum_{v=1}^{l}\|\mathbf{S}^{(v)}-\sum_{i=1, i \neq v}^{l} \mathbf{S}^{(i)} \mathbf{R}_{i v}\|_{F}^{2}+\|\mathbf{R}\|_{F}^{2} \\
&{ s.t.}~0 \leq \mathbf{S}^{(v)} \leq 1, \mathbf{S}^{(v) T} \mathbf{1}=\mathbf{1}, \mathbf{S}_{i i}^{(v)}=0, \\ & 0 \leq \mathbf{R}_{i v} \leq 1,\sum_{i=1, i \neq v}^{l} \mathbf{R}_{i v}=1, \mathbf{R}_{v v}=0,
\end{aligned}
\end{equation}
where $\mathbf{S}^{(v)} \in \mathbb{R}^{N \times N}$ is the similarity matrix of view $v$ and $\mathbf{R}\in \mathbb{R}^{l \times l}$ can be deemed as the  regression coefficient matrix. The Frobenius norm imposed on $\mathbf{R}$ is beneficial to improve robustness to noise~\cite{C.Y.Lu2012}. Compared with using the predefined similarity matrix in incomplete multi-view data, Eq.~(\ref{5}) can obtain the more reliable similarity matrix by learning the complementary information from other views adaptively.  

In addition, in order to exploit the consensus information across different views, we assume that each view shares the same clustering indicator matrix $\mathbf{V}$, i.e., $\mathbf{V}^{(1)}=\mathbf{V}^{(2)}=\cdots=\mathbf{V}^{(l)}=\mathbf{V}$ in Eq.~(\ref{NWNMF2}). Moreover,  according to the spectral graph theory~\cite{P.F.Zhu2017, Z.C.Li2015}, if two data points have high similarity in the intrinsic graph of data, then their corresponding clustering labels also have high similarity. Then, we can add the following regularization term to preserve the local geometric structure in each view.  
\begin{equation}\label{6}
	\begin{aligned}
		&\min _{\mathbf{V}} \sum_{v=1}^{l}\operatorname{Tr}(\mathbf{V}^{T}\mathbf{L}_{s}^{(v)}\mathbf{V})\\
		&{ s.t. } \mathbf{V} \geq 0,\mathbf{V}^{T}\mathbf{V}=\mathbf{I},
	\end{aligned}
\end{equation}
where $\mathbf{L}_{s}^{(v)}=\mathbf{D}_{s}^{(v)}-\mathbf{S}^{(v)}$ denotes the Laplacian matrix and $\mathbf{D}_{s}^{(v)}$ is  the diagonal matrix with $\mathbf{D}_{s}^{(v)}(i,i)=\sum_{j=1}^{N} \mathbf{S}^{(v)}_{ij}$. 

By combining Eqs.~(\ref{NWNMF2}),  (\ref{5}) and (\ref{6}) together,  the proposed incomplete multi-view unsupervised feature selection method (C$^{2}$IMUFS) is summarized as follows: 
\begin{equation}\label{7}
	\begin{aligned}
		\min _{\Omega} &\sum_{v=1}^{l}(\alpha^{(v)})^{\gamma}[\|(\mathbf{X}^{(v)}-\mathbf{U}^{(v)} \mathbf{V}^{T})\mathbf{W}^{(v)}\|_{F}^{2}+\lambda\|\mathbf{U}^{(v)}\|_{2,p}^{p}\\&+\!\beta (\operatorname{Tr}(\mathbf{V}^{T}\mathbf{L}_{s}^{(v)}\mathbf{V})\!+\!\|\mathbf{S}^{(v)}\!\!-\!\!\sum_{i=1, i \neq v}^{l} \mathbf{S}^{(i)} \mathbf{R}_{i v}\|_{F}^{2}\!+\!\|\mathbf{R}\|_{F}^{2})]\\
		{s.t.}~ &\mathbf{V} \geq 0,\mathbf{V}^{T}\mathbf{V}=\mathbf{I},\mathbf{U}^{(v)} \geq 0,0 \leq \alpha^{(v)} \leq 1,\sum_{v=1}^{l} \alpha^{(v)}=1 \\&0 \leq \mathbf{S}^{(v)} \leq 1,\mathbf{S}^{(v) T} \mathbf{1}=\mathbf{1}, \mathbf{S}_{i i}^{(v)}=0,0 \leq \mathbf{R}_{i v} \leq 1 \\& \sum_{i=1, i \neq v}^{l} \mathbf{R}_{i v}=1, \mathbf{R}_{v v}=0,v=1,2, \ldots, l,
	\end{aligned}
\end{equation}
where $\Omega=\{\mathbf{U}^{(v)},\mathbf{S}^{(v)},\alpha^{(v)},\mathbf{V},\mathbf{R} | v=1,2, \ldots, l\}$ and $\beta\ge 0$ is a trade-off hyper-parameter.

As can be seen from Eq.~(\ref{7}), the proposed method C$^{2}$IMUFS has two advantages: On one hand, the WNMF-based multi-view unsupervised feature selection model equipped with the adaptive view-weight learning and the $\ell_{2,p}$-norm regularization can better employ the available information of the incomplete data and automatically determine the view importance. It also provides a more flexible way to select informative features by adjusting $p$ to control the sparsity  according to the data.   On the other hand, C$^{2}$IMUFS can obtain the more precise similarity-induced graph matrix in each view and  preserve the reliable local  geometric structure  by simultaneously exploiting the complementary and consensus information among different views.

\section{Optimization and Algorithm}\label{sec:optimization}
Since the objective function in (\ref{7}) is not convex to all variables simultaneously, we propose an alternative iterative algorithm to solve the optimization problem, namely, optimizing the  objective function with regards to one variable while others fixed.
\subsection{Update $\mathbf{U}^{(v)}$ by Fixing Other Variables}
When fixing  other variables and removing the irrelevant terms, $\mathbf{U}^{(v)}$ can be updated by solving the following problem:
\begin{equation}\label{N7}
	\begin{aligned}
		\min _{\mathbf{U}^{(v)}} &(\alpha^{(v)})^{\gamma}[\|(\mathbf{X}^{(v)}-\mathbf{U}^{(v)} \mathbf{V}^{T})\mathbf{W}^{(v)}\|_{F}^{2}+\lambda\|\mathbf{U}^{(v)}\|_{2,p}^{p}]\\
		{s.t.}~ &\mathbf{U}^{(v)} \geq 0.
	\end{aligned}
\end{equation}
By introducing a Lagrange multiplier $\mathbf{\Psi}$  for the non-negative constraint $\mathbf{U}^{(v)}\ge 0$, we have the following Lagrange function:
\begin{equation}\label{11}
\begin{aligned}
\mathcal{L}(\mathbf{U}^{(v)})\!=&\!(\alpha^{(v)}\!)^{\gamma}\![\|(\mathbf{X}^{(v)}-\mathbf{U}^{(v)} \mathbf{V}^{T})\mathbf{W}^{(v)}\|_{F}^{2}+\lambda\|\mathbf{U}^{(v)}\|_{2,p}^{p}]\\&+Tr(\mathbf{\Psi}^{T}\mathbf{U}^{(v)}).
\end{aligned}
\end{equation}
In order to obtain the derivative of $\|\mathbf{U}^{(v)}\|_{2,p}^{p}$ w.r.t. $\mathbf{U}^{(v)}$, we introduce a lemma as follows:
\begin{lemma}\label{lemma1}
	\begin{align}
	&\lim _{\mathbf{u}^{i} \rightarrow \mathbf{u}_{*}^{i}} \frac{\partial\|\mathbf{U}\|_{2,p}^{p}}{\partial \mathbf{u}^{i}}=\lim _{\mathbf{u}^{i} \rightarrow \mathbf{u}_{*}^{i}} \frac{\partial(\left\|\mathbf{u}^{i}\right\|_{2})^{p}}{\partial \mathbf{u}^{i}}\notag\\
	& =\frac{p\mathbf{u}_{*}^{i}(\left\|\mathbf{u}_{*}^{i}\right\|_{2})^{p-1}}{\left\|\mathbf{u}_{*}^{i}\right\|_{2}} \\
	&\lim _{\mathbf{u}^{i} \rightarrow \mathbf{u}_{*}^{i},\epsilon \rightarrow 0} \frac{\partial\operatorname{Tr}(\mathbf{U}^{T}\mathbf{D}\mathbf{U})}{\partial \mathbf{u}^{i}}=\lim _{\mathbf{u}^{i} \rightarrow \mathbf{u}_{*}^{i},\epsilon \rightarrow 0} 2\mathbf{D}_{ii}\mathbf{u}^{i}\notag\\
	&=\frac{p\mathbf{u}_{*}^{i}(\left\|\mathbf{u}_{*}^{i}\right\|_{2})^{p-1}}{\left\|\mathbf{u}_{*}^{i}\right\|_{2}} ,
	\end{align}
	where $\mathbf{D}$ is a diagonal matrix and the corresponding diagonal entry $\mathbf{D}_{ii}=\frac{p(\|\mathbf{u}^{i}_{*}\|_{2})^{p-1}}{2\sqrt{\|\mathbf{u}^{i}_{*}\|_{2}^{2}+\epsilon}}$ and $\epsilon \rightarrow 0$ is a small value that prevents the denominator from being 0.
\end{lemma}

According to Lemma~\ref{lemma1}, the derivative of Eq.~(\ref{11}) with respect to $\mathbf{U}^{(v)}$ is equivalent to the derivative of the following function:
\begin{equation}\label{12}
	\begin{aligned}
		\mathcal{L}(\mathbf{U}^{(v)})=&(\alpha^{(v)})^{\gamma}[\|(\mathbf{X}^{(v)}-\mathbf{U}^{(v)} \mathbf{V}^{T})\mathbf{W}^{(v)}\|_{F}^{2}+\\&\lambda\operatorname{Tr}(\mathbf{U}^{(v)T}\mathbf{D}^{(v)}\mathbf{U}^{(v)})]+Tr(\mathbf{\Psi}^{T}\mathbf{U}^{(v)}).
	\end{aligned}
\end{equation}

Then, by taking the partial derivative of $\mathcal{L}(\mathbf{U}^{(v)})$ w.r.t. $\mathbf{U}^{(v)}$ in Eq.~(\ref{12}), we have
\begin{equation}\label{14}
\begin{aligned}
\frac{\partial\mathcal{L}(\mathbf{U}^{(v)})}{\partial\mathbf{U}^{(v)}}&=(\alpha^{(v)})^{\gamma}[2(\mathbf{U}^{(v)}\mathbf{V}^{T}\mathbf{W}^{(v)}\mathbf{W}^{(v)}\mathbf{V}-\\
&\mathbf{X}^{(v)}\mathbf{W}^{(v)}\mathbf{W}^{(v)}\mathbf{V})+2\lambda\mathbf{D}^{(v)}\mathbf{U}^{(v)}]+\mathbf{\Psi}.
\end{aligned}
\end{equation}
According to the Karush-Kuhn-Tucker (KKT)  complementary condition $\mathbf{\Psi}_{ij}\mathbf{U}^{(v)}_{ij}=0$~\cite{KKT}, $\mathbf{U}^{(v)}$ can be updated as follows:
\begin{equation}\label{15}
\mathbf{U}^{(v)}_{ij} \!\leftarrow\!\! \mathbf{U}^{(v)}_{ij}\!\!\sqrt{\!\!\frac{(\mathbf{X}^{(v)}\mathbf{W}^{(v)}\mathbf{W}^{(v)}\mathbf{V})_{ij}}{(\mathbf{U}^{(v)}\mathbf{V}^{T}\mathbf{W}^{(v)}\mathbf{W}^{(v)}\mathbf{V}+\lambda\mathbf{D}^{(v)}\mathbf{U}^{(v)})_{ij}}}.
\end{equation}

\subsection{Update $\mathbf{V}$ by Fixing Other Variables}
When other variables are fixed, the problem of solving $\mathbf{V}$ is degraded to minimize the following optimization problem by removing irrelevant terms:
\begin{equation}\label{updateV}
	\begin{aligned}
		\min_{\mathbf{V}}&\!\!\sum_{v=1}^{l}\!(\!\alpha^{(v)}\!)\!^{\gamma}\![\|(\mathbf{X}^{(v)}-\mathbf{U}^{(v)} \mathbf{V}^{T})\mathbf{W}^{(v)}\|_{F}^{2}+\beta (\operatorname{Tr}(\mathbf{V}^{T}\mathbf{L}_{s}^{(v)}\mathbf{V})\!]\!\\
		{s.t.}~ &\mathbf{V} \geq 0,\mathbf{V}^{T}\mathbf{V}=\mathbf{I}.
	\end{aligned}
\end{equation}

Then, we can construct the following Lagrange function by introducing the Lagrange multipliers $\xi$ and $\mathbf{\Theta}$ for the orthogonal constraint $\mathbf{V}^{T}\mathbf{V}=\mathbf{I}$ and the non-negative constraint $\mathbf{V} \geq 0$, respectively.
\begin{equation}\label{8}
	\begin{aligned}
		&\mathcal{L}(\mathbf{V})=\sum_{v=1}^{l}(\alpha^{(v)})^{\gamma}\|(\mathbf{X}^{(v)}-\mathbf{U}^{(v)} \mathbf{V}^{T})\mathbf{W}^{(v)}\|_{F}^{2} \\&+\beta\operatorname{Tr}(\mathbf{V}^{T}\mathbf{L}_{s}^{(v)}\mathbf{V})]+\xi\|\mathbf{V}^{T}\mathbf{V}-\mathbf{I}\|_{F}^{2}+\operatorname{Tr}(\mathbf{\Theta}^{T}\mathbf{V}).\\
	\end{aligned}
\end{equation}
To guarantee the orthogonality of $\mathbf{V}$,  $\xi$ is set to a large constant ($\xi=10^{7}$ in our experiments).

By taking the derivative of $\mathcal{L}(\mathbf{V})$ w.r.t. $\mathbf{V}$, we have
\begin{equation}\label{9}
	\begin{aligned}
		&\frac{\partial\mathcal{L}(\mathbf{V})}{\partial\mathbf{V}}=\sum_{v=1}^{l}(\alpha^{(v)})^{\gamma}(2\mathbf{W}^{(v)}\mathbf{W}^{(v)}\mathbf{V}\mathbf{U}^{(v)T}\mathbf{U}^{(v)}+2\beta\mathbf{L}_{s}^{(v)}\mathbf{V}\\&-2\mathbf{W}^{(v)}\mathbf{W}^{(v)}\mathbf{X}^{(v)T}\mathbf{U}^{(v)}+\xi(4\mathbf{V}\mathbf{V}^{T}\mathbf{V}-4\mathbf{V}))+\mathbf{\Theta}.
	\end{aligned}
\end{equation}

Based on the KKT complementary condition $\mathbf{\Theta}_{ij}\mathbf{V}_{ij}=0$, we obtain the updating rule for $\mathbf{V}$ as follows:
\begin{equation}\label{10}
	\mathbf{V}_{ij} \leftarrow \mathbf{V}_{ij}\sqrt{\frac{(\mathbf{E}^{(v)}+2\xi\mathbf{V})_{ij}}{(\mathbf{Q}^{(v)}+2\xi\mathbf{V}\mathbf{V}^{T}\mathbf{V})_{ij}}},
\end{equation}
where $\mathbf{Q}^{(v)}=\sum_{v=1}^{l}(\alpha^{(v)})^{\gamma}(\mathbf{W}^{(v)}\mathbf{W}^{(v)}\mathbf{V}\mathbf{U}^{(v)T}\mathbf{U}^{(v)}+\beta\mathbf{D}_{s}^{(v)}\mathbf{V})$, $\mathbf{E}^{(v)}=\sum_{v=1}^{l}(\alpha^{(v)})^{\gamma}(\mathbf{W}^{(v)}\mathbf{W}^{(v)}\mathbf{X}^{(v)T}\mathbf{U}^{(v)}+\beta\mathbf{S}^{(v)}\mathbf{V})$.

\subsection{Update $\mathbf{S}^{(v)}$ by Fixing Other Variables}
With other variables fixed, we can update $\mathbf{S}^{(v)}$ by solving the following problem:
\begin{equation}\label{16}
\begin{aligned}
&\min _{\mathbf{S}^{(v)}} \sum_{v=1}^{l} (\alpha^{(v)})^{r}(\operatorname{Tr}(\mathbf{V}^{T}\mathbf{L}_{s}^{(v)}\mathbf{V})+\|\mathbf{S}^{(v)}-\sum_{i=1, i \neq v}^{l} \mathbf{S}^{(i)} \mathbf{R}_{i v}\|_{F}^{2})\\
&{s.t.}~0 \leq \mathbf{S}^{(v)} \leq 1, \mathbf{S}^{(v) T} \mathbf{1}=\mathbf{1}, \mathbf{S}_{i i}^{(v)}=0.
\end{aligned}
\end{equation}
We first introduce some notations for the convenience of following discussion. Let $\mathbf{N}^{(i)}=\mathbf{S}^{(i)}-\sum_{j=1, j \neq v, j \neq i}^{l} \mathbf{S}^{(j)} \mathbf{R}_{ji}$, $\mathbf{B}^{(v)}=\sum_{i=1,i\neq v}^{l} \mathbf{S}^{(i)} \mathbf{R}_{iv}$ and $\mathbf{H}_{ij}=\|\mathbf{v}^{i}-\mathbf{v}^{j}\|_{2}^{2}$. Then, Eq.~(\ref{16}) can be rewritten as follows:
\begin{equation}\label{17}
\begin{aligned}
& \min _{0 \leq \mathbf{S}^{(v)} \leq 1,\mathbf{S}^{(v) T} \mathbf{1}=\mathbf{1},\mathbf{S}_{i i}^{(v)}=0} (\alpha^{(v)})^{\gamma}\operatorname{Tr}(\mathbf{V}^{T}\mathbf{L}_{s}^{(v)}\mathbf{V})+\\
&\sum_{v=1}^{l} (\alpha^{(v)})^{\gamma}\|\mathbf{S}^{(v)}-\sum_{i=1, i \neq v}^{l} \mathbf{S}^{(i)} \mathbf{R}_{i v}\|_{F}^{2}\\
\Leftrightarrow & \min _{0 \leq \mathbf{S}^{(v)} \leq 1,\mathbf{S}^{(v) T} \mathbf{1}=\mathbf{1},\mathbf{S}_{i i}^{(v)}=0} \frac{(\alpha^{(v)})^{\gamma}}{2}\sum_{i,j}^{n}\|\mathbf{v}^{i}-\mathbf{v}^{j}\|_{2}^{2}\mathbf{S}_{ij}^{(v)}+\\
&\sum_{i=1, i \neq v}^{l}(\alpha^{(i)})^{\gamma}\|\mathbf{R}_{vi}\mathbf{S}^{(v)}-(\mathbf{S}^{(i)}-\sum_{j=1, j \neq v,j \neq i}^{l}\mathbf{S}^{(j)}\mathbf{R}_{ji})\|_{F}^{2}\\
&+(\alpha^{(v)})^{\gamma}\|\mathbf{S}^{(v)}-\sum_{i=1, i \neq v}^{l}\mathbf{S}^{(i)}\mathbf{R}_{iv}\|_{F}^{2}\\
\Leftrightarrow & \min _{0 \leq \mathbf{S}^{(v)} \leq 1,\mathbf{S}^{(v) T} \mathbf{1}=\mathbf{1},\mathbf{S}_{i i}^{(v)}=0} \frac{(\alpha^{(v)})^{\gamma}}{2}\sum_{i,j}^{n}\mathbf{H}_{ij}\mathbf{S}_{ij}^{(v)}+\\
&\sum_{i=1, i \neq v}^{l}(\alpha^{(i)})^{\gamma}(\mathbf{R}_{vi})^{2}\|\mathbf{S}^{(v)}-\frac{1}{\mathbf{R}_{vi}}\mathbf{N}^{(i)}\|_{F}^{2}+\\
&(\alpha^{(v)})^{\gamma}\|\mathbf{S}^{(v)}-\mathbf{B}^{(v)}\|_{F}^{2}\\
\Leftrightarrow &\min _{0 \leq \mathbf{S}^{(v)} \leq 1,\mathbf{S}^{(v) T} \mathbf{1}=\mathbf{1},\mathbf{S}_{i i}^{(v)}=0} \sum_{i,j=1}^{N}(\mathbf{S}_{ij}^{(v)}-\mathbf{P}_{ij}^{(v)})^{2},
\end{aligned}
\end{equation}
where $\mathbf{P}_{ij}^{(v)}=\frac{\mathbf{Q}_{ij}^{(v)}}{\sum_{k=1, k \neq v}^{l}(\alpha^{(k)})^{\gamma}(\mathbf{R}_{vk})^{2}+(\alpha^{(v)})^{\gamma}}$ and $\mathbf{Q}_{ij}^{(v)}=\sum_{k=1, k \neq v}^{l}(\alpha^{(k)})^{\gamma}\mathbf{R}_{vk}\mathbf{N}_{ij}^{(k)}+(\alpha^{(v)})^{\gamma}(\mathbf{B}_{ij}^{(v)}-\frac{1}{4}\mathbf{H}_{ij})$. We can observe that all columns of  $\mathbf{S}^{(v)}$ are independent in problem (\ref{17}). Then, we can obtain the following optimal solution of problem (\ref{16}) according to \cite{F.P.Nie2016,J.Wen2020}:
\begin{equation}\label{18}
\mathbf{S}_{ij}^{(v)}=\left\{\begin{array}{cc}
(\mathbf{P}_{ij}^{(v)}+\delta_{i})_{+} & i \neq j; \\
0 & i=j,
\end{array}\right.
\end{equation}
where $\delta_{i}=(1-\sum_{j=1, i \neq j}^{n} \mathbf{P}_{ji}^{(v)}) /(n-1)$.
\subsection{Update $\mathbf{R}$ by Fixing Other Variables}
When other variables are fixed and the irrelevant terms are removed, updating $\mathbf{R}$ is equal to solve the following problem:
\begin{equation}\label{19}
	\begin{aligned}
		&\min _{\mathbf{R}} \sum_{v=1}^{l}\|\mathbf{S}^{(v)}-\sum_{i=1, i \neq v}^{l} \mathbf{S}^{(i)} \mathbf{R}_{i v}\|_{F}^{2}+\|\mathbf{R}\|_{F}^{2}\\
		&{s.t.}~0 \leq \mathbf{R}_{i v} \leq 1, \sum_{i=1, i \neq v}^{l} \mathbf{R}_{i v}=1, \mathbf{R}_{v v}=0.
	\end{aligned}
\end{equation}

Let $\mathbf{K}=[Vec(\mathbf{S}^{(1)}),Vec(\mathbf{S}^{(2)}),\cdots,Vec(\mathbf{S}^{(l)})] \in \mathbb{R}^{N^{2} \times l}$, where $Vec(\cdot)$ indicates a vec-operator stacking each column of $\mathbf{S}^{(v)}~(v=1,2,\cdots, l)$ on top of the next. Then, problem (\ref{19}) can be transformed as follows:
\begin{equation}\label{20}
	\begin{aligned}
		&\min _{\mathbf{R}} \sum_{v=1}^{l}(\|\mathbf{K}(:,v)-\mathbf{K}\mathbf{R}(:,v)\|_{2}^{2}+\|\mathbf{R}(:,v)\|_{2}^{2}) \\
		&{s.t.}~0 \leq \mathbf{R}_{i v} \leq 1, \sum_{i=1, i \neq v}^{l} \mathbf{R}_{i v}=1, \mathbf{R}_{v v}=0.
	\end{aligned}
\end{equation}
where $\mathbf{K}(:,v)$ and $\mathbf{R}(:,v)$ denote the $v$-th column of $\mathbf{K}$ and $\mathbf{R}$, respectively. As seen from Eq.~(\ref{20}), it can be divided into $l$ independent sub-problems. Then, we can obtain the following sub-problem w.r.t. $\mathbf{R}(:,v)$:
\begin{equation}\label{21}
\min _{0 \leq \mathbf{R}_{i v} \leq 1,\sum_{i=1, i \neq v}^{l} \mathbf{R}_{i v}=1, \mathbf{R}_{v v}=0} \|\mathbf{K}\!(:,v\!)\!-\!\mathbf{K}\mathbf{R}\!(:,v\!)\|_{2}^{2}\!+\!\|\mathbf{R}\!(:,v\!)\|_{2}^{2}.
\end{equation}
Problem (\ref{21}) can be solved by the accelerated projected gradient algorithm in~\cite{J.Huang2015}.

\subsection{Update $\alpha^{(v)}$ by Fixing Other Variables}
To update $\alpha^{(v)}$ with other variables fixed, we first let $d^{(v)}=\|(\mathbf{X}^{(v)}-\mathbf{U}^{(v)} \mathbf{V}^{T})\mathbf{W}^{(v)}\|_{F}^{2}+\lambda\|\mathbf{U}^{(v)}\|_{2,p}^{p}+\beta (\operatorname{Tr}(\mathbf{V}^{T}\mathbf{L}_{s}^{(v)}\mathbf{V})+\|\mathbf{S}^{(v)}-\sum_{i=1, i \neq v}^{l} \mathbf{S}^{(i)} \mathbf{R}_{i v}\|_{F}^{2})$. Then, the problem of solving  $\alpha^{(v)}$ is reduced as follows:
\begin{equation}\label{n77}
	\min _{\alpha^{(v)}} \sum_{v=1}^{l}(\alpha^{(v)})^{\gamma}d^{(v)}, s.t. \sum_{v=1}^{l} \alpha^{(v)}=1, 0 \leq \alpha^{(v)} \leq 1.
\end{equation}
We introduce the following Lagrange function to solve  $\alpha^{(v)}$. 
\begin{equation}\label{22}
\mathcal{L}(\alpha^{(v)})=\sum_{v=1}^{l}(\alpha^{(v)})^{\gamma} d^{(v)}-\psi(\sum_{v=1}^{l} \alpha^{(v)}-1),
\end{equation}
where $\psi$ is the Lagrange multiplier.
Taking the partial derivative of $\mathcal{L}$ w.r.t. $\alpha^{(v)}$ and setting it to zero, we have
\begin{equation}\label{23}
\begin{aligned}
&\frac{\partial \mathcal{L}(\alpha^{(v)})}{\partial \alpha^{(v)}}=\gamma\left(\alpha^{(v)}\right)^{r-1} d^{(v)}-\psi=0\\
\Rightarrow&\quad\alpha^{(v)} = (\frac{\psi}{\gamma d^{(v)}})^{1/(\gamma-1)},
\end{aligned}
\end{equation}
Besides, since $\sum_{v=1}^{l}\alpha^{(v)}=1$, we can get the final solution of $\alpha^{(v)}$ as follows:
\begin{equation}\label{24}
\alpha^{(v)}=(\frac{d^{(v)}}{\sum_{v=1}^{l} (d^{(v)})})^{1 /(1-\gamma)}.
\end{equation}
The overall optimization procedure of C$^{2}$IMUFS is summarized in Algorithm 1.

\begin{algorithm}
\caption{Iterative algorithm of C$^{2}$IMUFS}
\KwIn{ \begin{enumerate}
\item The incomplete multi-view data $\mathcal{X}=\{\mathbf{X}^{(v)}\}_{v=1}^{l}$;
\item The weight matrices $\{\mathbf{W}^{(v)}\}_{v=1}^{l}$;
\item The selected features $h$;
\item The parameters $\lambda$, $\beta$ and $\gamma$.
\end{enumerate}
}

$\mathbf{Initialize:}$ $\{\mathbf{U}^{(v)}\in \mathbb{R}^{d^{v}\times c} \}_{v=1}^{l}$, $\mathbf{V}\in \mathbb{R}^{N \times c}$, $\{\mathbf{S}^{(v)} \in \mathbb{R}^{N\times N} \}_{v=1}^{l}$ and $\{\alpha^{(v)}=1 / l\}_{v=1}^{l}$.\\

\Begin
{
\While{not convergent}{
\quad Update $\mathbf{V}$ via (\ref{10});\\
\quad Update $\left\{\mathbf{U}^{(v)}\right\}_{v=1}^{l}$ via (\ref{15});\\
\quad Update $\left\{\mathbf{S}^{(v)}\right\}_{v=1}^{l}$ via (\ref{18});\\
\quad Update $\mathbf{R}$ by solving (\ref{21});\\
\quad Update $\left\{\alpha^{(v)}\right\}_{v=1}^{l}$ via (\ref{24}).\\
}
}
\KwOut{Sorting the $\ell_{2}$-norm of the rows of $\{\mathbf{U}^{(v)}\}_{v=1}^{l}$ in a descending order and selecting the top $h$ features.}
\end{algorithm}

\section{Convergence and complexity analysis}\label{sec:Discussions}
In this section, we give the theoretical analysis on convergence  and complexity of the proposed algorithm  C$^{2}$IMUFS.

\subsection{Convergence Analysis}\label{sec:convergence}
Since the objective function in Eq.~(\ref{7}) is not convex for the variables $\mathbf{U}^{(v)},\mathbf{V}, \mathbf{S}^{(v)}, \mathbf{R}$ and $\alpha^{(v)}$ simultaneously, we divide (\ref{7}) into five sub-objective functions, namely, (\ref{N7}), (\ref{updateV}), (\ref{16}),  (\ref{19}) and (\ref{n77}).  Hence, we can prove the convergence of Algorithm 1 by proving the monotonic convergence of  each sub-objective functions. We first introduce the following theorem to prove the decrease of the objective function in problem~(\ref{N7}) while updating $\mathbf{U}^{(v)}$ with other variables fixed. 

\begin{theorem}\label{Theorem2}
	By using the updating rules in Algorithm 1, the objective function in Eq. (\ref{N7}) monotonically decreases until convergence.
\end{theorem}

We can prove this theorem according to the  definition and lemma proposed in~\cite{C. Ding2008}. Due to the limited space, the detailed proof of  Theorem 1 can be found in the supplementary material. 

We can prove the monotonic convergence of updating $\mathbf{V}$ in a similar way. In addition, the convergences of updating variables $\mathbf{S}^{(v)}$ and $\alpha^{(v)}$ can be guaranteed by their closed solutions in Eqs.~(\ref{18}) and (\ref{24}), respectively. Besides,  the convergence of $\mathbf{R}$ can be obtained according to~\cite{J.Huang2015}. Hence, the objective function in Eq.~(\ref{7}) will converge according to the updating rules in (\ref{15}),  (\ref{10}), (\ref{18}), (\ref{21}) and (\ref{24}). Furthermore, we will verify the convergence behavior of Algorithm 1 in the experiment section. 

\subsection{Complexity Analysis}
In Algorithm 1, there are five variables need to be updated. For updating $\mathbf{U}^{(v)}$, the computational complexity is dominated by the matrix multiplication operation, which has a complexity of $\mathcal{O}(Nd_{v}c)$. Similar to updating  $\mathbf{U}^{(v)}$, the computational complexity of updating $\mathbf{V}$ is  $\mathcal{O}(Nmax(d_{v},N)c)$. For updating $\mathbf{S}^{(v)}$, the computational cost only consists of some  element-based operations, which can be ignored. For updating $\mathbf{R}$, a projected gradient algorithm proposed in \cite{J.Huang2015} is employed to solve problem~(\ref{21}). According to \cite{J.Huang2015}, the projected gradient algorithm also comprises  some element-based operations. Thus, the computational complexity of updating $\mathbf{R}$ is also ignored. As to updating  $\alpha^{(v)}$, it only involves the numerical division and addition operations in terms of Eq.~(\ref{24}). Hence, the total time complexity of Algorithm 1 is $\mathcal{O}(Nmax(d_{v},N)ct)$, where $t$ is the number of iteration.

\section{Experiments}\label{sec:Experimental}
In this section, we demonstrate the effectiveness of our method by comparing with several state-of-the-art unsupervised feature selection methods in terms of clustering performance. 

\subsection{Experimental Schemes}
\subsubsection{Datasets}
In experiments, eight real-world multi-view datasets are used to investigate the performance of our method. The descriptions of these datasets are presented as follows.

BBCSport\footnote{http://mlg.ucd.ie/datasets/segment.html}: It is a multi-view document dataset collected from the BBC Sport website. We follow  the experimental settings in \cite{bbcs}, where  116 news objects with four views are selected from the entire datasets.

3Sources\footnote{http://mlg.ucd.ie/datasets/3sources.html}: It  contains 169 news objects with six topics, which are collected from three online news medias. 

Washington Dataset\footnote{http://elenaher.dinauz.org/phd/data/networks/webkb/}:  It is one of the famous WebKB dataset, which contains 203 web-pages described by three views including the content features, title features and citation features. 

BBC4view$^{1}$: It consists of  685 news articles characterized by four views. These articles are divided into five topics including  politics, entertainment, business,  sports and technology.

Reuters\footnote{http://www.research.att.com/lewis}: It contains 1200 documents, which are characterized by five different languages (views).

Caltech101-7\footnote{http://www.vision.caltech.edu/Image\_Datasets/Caltech101/}: It is a multi-view image dataset containing 1474 pictures of objects. Each image are characterized by  six visual features. 

Handwritten 2 sources (HW2sources)~\cite{HW2}: It contains 2000 objects of ten handwritten digits (0-9) derived from two different information sources, namely, MNIST and USPS.

CiteSeer~\cite{CiteSeer}: It is a collection of scientific publications, which contains 3312 documents classified into six categories. 

The detailed statistics of these multi-view datasets are summarized in Table~\ref{Table1}. Besides, we  simulate the incomplete multi-view setting according to the proposed method in \cite{X.W.Liu2018}, namely, removing  certain ratios of instances from each view as missing data. The missing ratio is set from 10\% to 50\% with a step of 10\%.

\begin{table}[htbp]\footnotesize
\tabcolsep 0pt
\caption{A detail description of datasets}\label{Table1}
\vspace*{-15pt}
\renewcommand\tabcolsep{1.5pt} 
    \begin{flushleft}
    \def\temptablewidth{\textwidth}
        \resizebox{8.7cm}{!}{
        \begin{tabular}{@{\extracolsep{\fill}}lccccc}
            \toprule
            Datasets  &Abbr.& Views& Instances & Features & Classes\\
            \hline
                BBCSport& BBCS       & 4 & 116 &1991/2063/2113/2158 &  5\\
              3Sources &3S         & 3 & 169 &3560/3631/3068& 6 \\
              Washington&Washington     & 3 & 203&1703/230/230& 4 \\
              BBC4view& BBC4        &4   &685  &4659/4633/4665/4684  &5\\
              Reuters &Reu &5 &1200 &2000/2000/2000/2000/2000&6\\
              Caltech101-7&Cal7          & 6 & 1474&48/40/254/1984/512/928&  7  \\
              HW2sources&HW2    & 2 & 2000&784/256& 10\\
              CiteSeer &CS &2 &3312 &3312/3703 &6\\
              \hline
        \end{tabular}}
    \end{flushleft}
\end{table}

\subsubsection{Evaluation metrics}
Two widely used clustering metrics including clustering accuracy (ACC) and normalized mutual information (NMI) are employed to measure the performance of  compared methods. They can be defined as follows, respectively.  
\begin{equation}\label{37}
	\mathrm{ACC}=\frac{1}{n} \sum_{i=1}^{n} \delta\left(y_{i}, \operatorname{map}\left(\hat{y_{i}}\right)\right),
\end{equation}
where $y_{i}$ and $\hat{y_{i}}$ denote the ground truth label and clustering result of the $i$-th instance, respectively, $n$ is the number of samples, $\delta(x,y)$ is an indicator function satisfying $\delta(x,y)=1$ for $x=y$, otherwise $\delta(x,y)=0$, and $\operatorname{map}(\cdot)$ denotes the permutation mapping function which maps each cluster index to the best ground true label using Kuhn-Munkres algorithm\cite{A.Strehl2002}.
\begin{equation}\label{38}
	\operatorname{NMI}\left(\mathcal{Y}, \mathcal{Y}^{\prime}\right)=\frac{\operatorname{MI}\left(\mathcal{Y}, \mathcal{Y}^{\prime}\right)}{\max \left(H(\mathcal{Y}), H\left(\mathcal{Y}^{\prime}\right)\right)},
\end{equation}
where $\mathcal{Y}$ and $\mathcal{Y}^{\prime}$ represent the set of true labels and clustering labels respectively, $\operatorname{MI}(\mathcal{Y},\mathcal{Y}^{\prime})$ is the mutual information between $\mathcal{Y}$ and $\mathcal{Y}^{\prime}$, and $H(\mathcal{Y})$ and $H(\mathcal{Y}^{\prime})$ are the entropies of $\mathcal{Y}$ and $\mathcal{Y}^{\prime}$, respectively.

For these two metrics, the larger value indicates the better performance. 

\subsubsection{Comparison methods}
We compare our method C$^{2}$IMUFS with several state-of-the-art unsupervised feature selection methods including  four single-view based methods (i.e., LPscore, EGCFS, UDPFS and RUSLP) and six multi-view based methods (i.e., OMVFS,  ACSL, CGMV-UFS, NSGL, CVFS and CvLP-DCL). These compared methods are briefly introduced as follows: 

$\bullet$ \textbf{AllFea} uses all original features to be compared with. 

$\bullet$ \textbf{LPscore}~\cite{LapScore} selects features based on the Laplacian score, which measures the locality preserving power. 

$\bullet$ \textbf{EGCFS}~\cite{EGCFS}  incorporates the procedure of maximizing the between-class scatter matrix and the adaptive graph learning into a joint framework to select discriminative features.

$\bullet$ \textbf{UDPFS}~\cite{UDPFS} learns an unsupervised discriminative project for feature selection by introducing the fuzziness learning and sparse learning simultaneously. 

$\bullet$ \textbf{RUSLP}~\cite{RUSLP}  selects features by using the non-negative matrix factorization with  $\ell_{2}$-norm regularization and the preservation of local geometric structure. 

$\bullet$  \textbf{OMVFS}~\cite{OMVFS} designs an incremental non-negative matrix factorization model to select features. 

$\bullet$ \textbf{ACSL}~\cite{ACSL} incorporates the adaptive collaborative similarity structure learning and  multi-view feature selection into a unified framework. 

$\bullet$ \textbf{CGMV-UFS}~\cite{CGMV-UFS} learns a consensus cluster indicator matrix among different views and embeds the feature selection into a non-negative matrix factorization framework. 

$\bullet$ \textbf{NSGL}~\cite{NSGL} integrates the adaptive similarity graph learning and multi-view feature selection into a unified framework by employing  the complementary information across views.

$\bullet$ \textbf{CVFS}~\cite{CVFS}  embeds the Hilbert–Schmidt Independence Criterion into a non-negative matrix factorization model for feature selection of streaming data. 

$\bullet$ \textbf{CvLP-DCL}~\cite{CvLP-DCL}  constructs two  modules including consensus learning and view-specific local structure learning to select discriminative features. 

We use these single-view based methods to select features by combining all features derived from different views into a single view. These parameters of compared methods are set according to the original papers for the guarantee of obtaining the optimal results. Meanwhile, the parameters $\lambda$ and $\beta$ in our method are tuned by searching the grid $\{10^{-3}, 10^{-2}, 10^{-1}, 1, 10, 10^{2}, 10^{3}\}$, $\gamma$ is tuned from $\{2, 3, 4, 5, 6, 7, 8\}$ and $p$  is tuned from $\{0.001,0.01,0.1,1\}$. Since  the optimal number of selected features is hard to determine in each dataset, we set the percentage of selected features from 10\% to 50\%  with an interval of 10\% for all datasets. Then, we run the incomplete multi-view clustering algorithm proposed in~\cite{cluster} 30 times on selected features and report the average results with standard deviation.

\subsection{Comparisons over Clustering Performance}
In this section, we compare the clustering performance of our method with other competing methods in terms of ACC and NMI. Tables~\ref{Table2} and~\ref{Table3} respectively show the clustering results ACC and NMI of different methods on eight datasets, where the best performance is highlighted with bold-face type. In addition, two-sample t-test is used to verify whether our proposed method C$^{2}$IMUFS is significantly superior to other methods. The values with $\bullet/ \circ$ in Tables~\ref{Table2} and~\ref{Table3} indicate that whether C$^{2}$IMUFS is statically better$/$inferior than other compared methods with the significance level of 0.05. 

From Tables~\ref{Table2} and~\ref{Table3}, we can see that our method C$^{2}$IMUFS is better than other methods in most cases. As to BBCS, Reu, BBC4 and HW2, C$^{2}$IMUFS gains over 13\% and  11\%  average improvement compared to all other methods in terms of ACC and NMI, respectively. As to 3S and CS, C$^{2}$IMUFS also obtains more than 5\% improvement in average ACC and NMI. On the Cal7 dataset, C$^{2}$IMUFS is still superior to the baseline in terms of both ACC and NMI. On the Washington dataset, C$^{2}$IMUFS outperforms other methods in ACC and gets the second best performance in NMI. In addition, when comparing with  the baseline method AllFea,  C$^{2}$IMUFS outperforms AllFea over all datasets  both in ACC and NMI, which demonstrates the effectiveness of the proposed method. Moreover, C$^{2}$IMUFS entirely outperforms all single-view based methods and achieves almost 10\% average improvement on most of datasets in terms of ACC and NMI. This demonstrates the effectiveness of our method using the consensus and complementary information among different views in comparison with these single-view based methods combining each data view by stacking.

\begin{figure*}[!htbp]
\centering
\includegraphics[width=\textwidth]{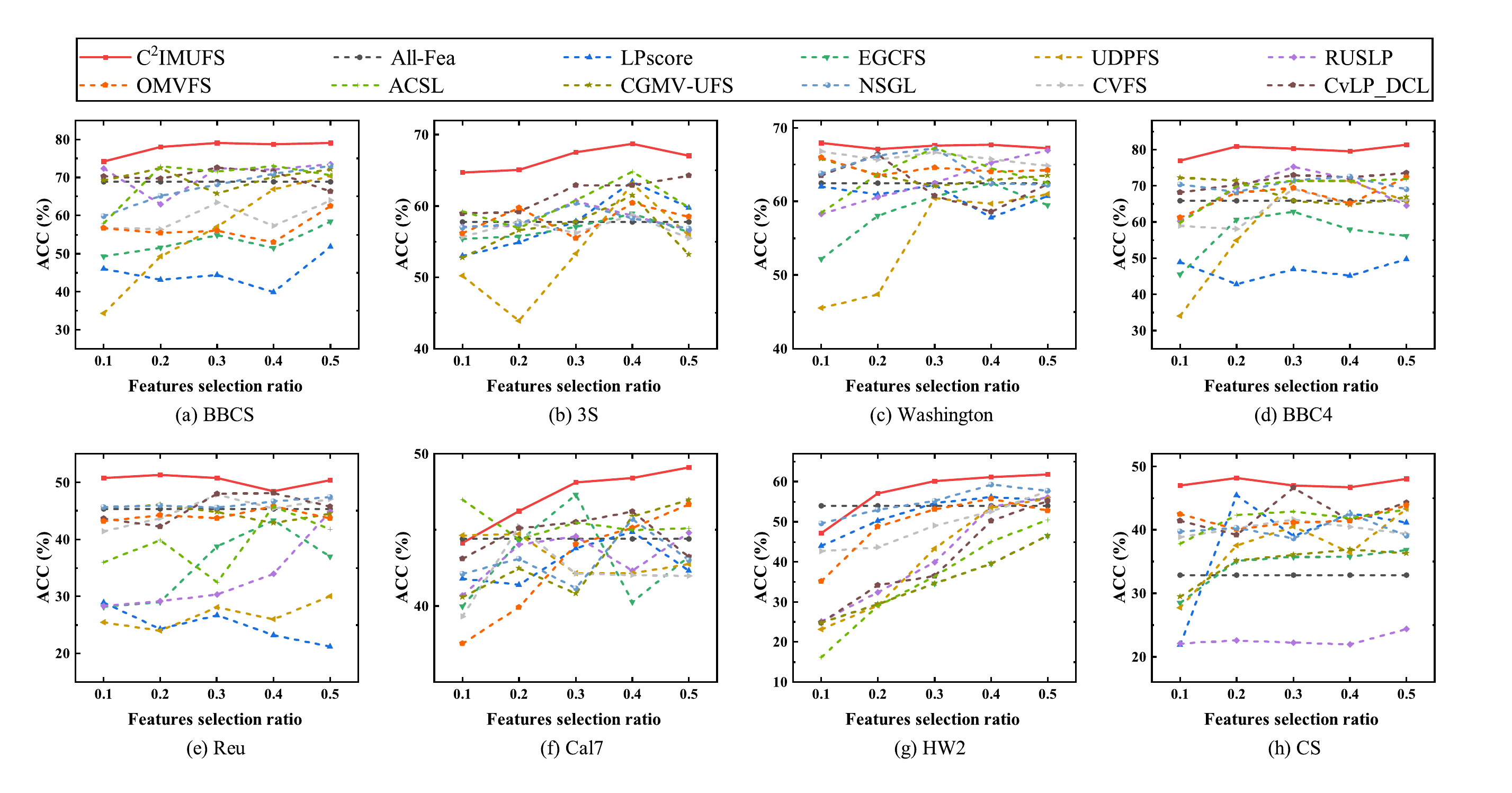}
\caption{ACC of different methods with different selected features ratios.}\label{ACC-missing0.3}
\end{figure*}

\begin{figure*}[!htbp]
\centering
\includegraphics[width=\textwidth]{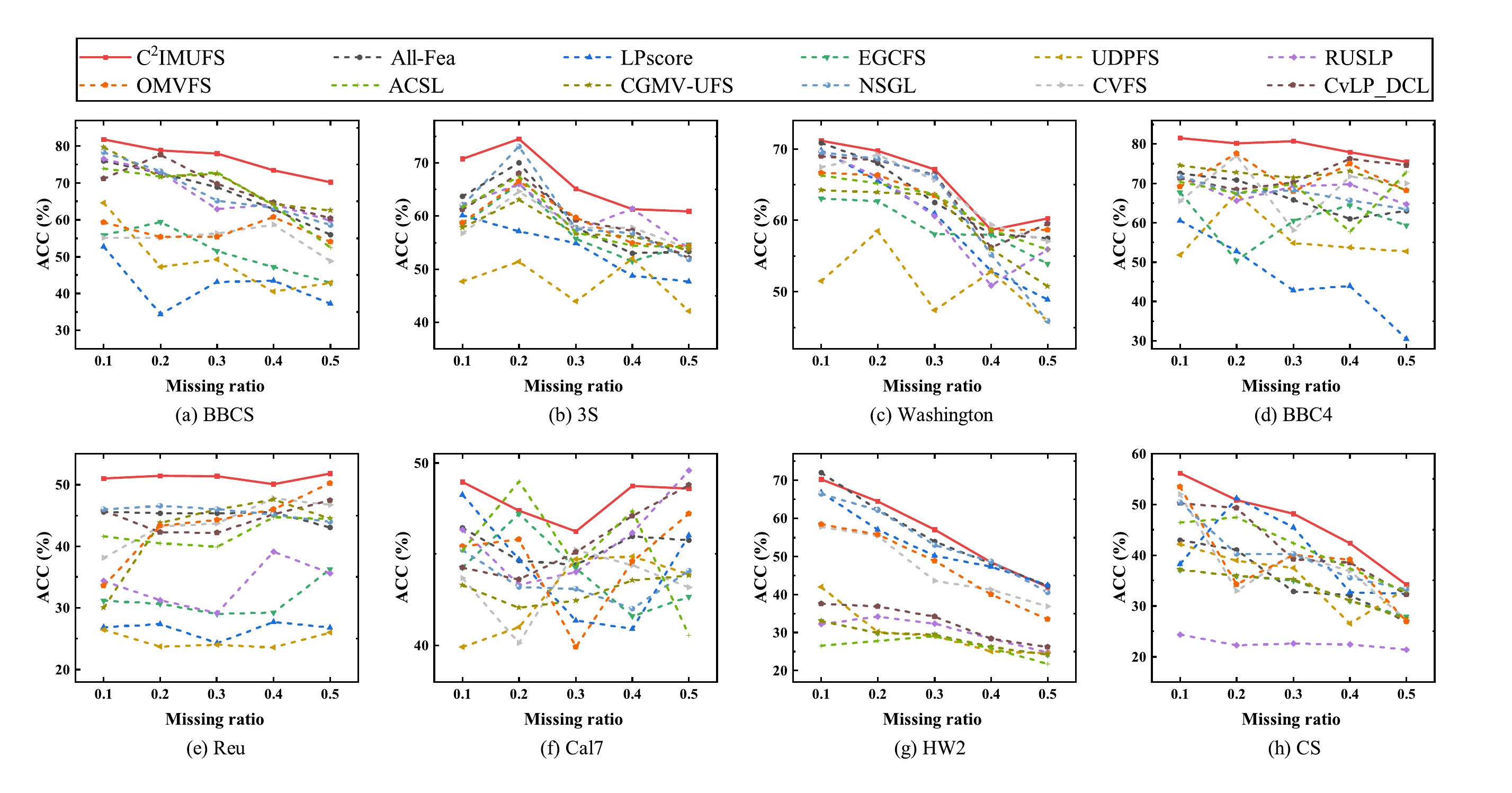}
\caption{ACC of different methods with different missing ratios.}\label{ACC-feature-selection-rate0.2}
\end{figure*}

Since it is hard to determine the optimal number of selected features in each dataset, we also show the clustering performances of different methods vary with the number of selected features. Due to the page limitation, we only report the experimental results of ACC. Fig.~\ref{ACC-missing0.3} plots the ACC values with different ratios of selected features while fixing the missing ratio at 0.3.  As can be seen, C$^{2}$IMUFS outperforms other methods with the variation of feature selection ratio from $10\%$ to $50\%$ in most cases. Furthermore, we also show the ACC of different methods with variation of missing ratio while the  feature selection ratio is set to $20\%$. From Fig.~\ref{ACC-feature-selection-rate0.2}, we can observe that C$^{2}$IMUFS is still better than other compared methods on all datasets in most of the time.  The experimental results of NMI with regards of different selected feature ratios and missing ratios can be founded in Figs. 1 and 2 in the supplementary material, respectively. In these two figures, we also can find that our method is still better than other methods in most of the cases. The superior performance of the proposed method is attributed to simultaneously learn the complementary and consensus information across different views and embed them into an extended WNMF framework with adaptive view-weight learning and a flexible sparse constraint.

\begin{table*}[!htbp]\huge 
	\tabcolsep 0pt
	\caption{Means and standard deviation (\%) of ACC for different methods on eight datasets with missing ratio 0.3 while selecting 20\% of all features.}\small\label{Table2}
	\vspace*{-5pt}
	\begin{flushleft}
		\def\temptablewidth{\textwidth}
		{\rule{\temptablewidth}{1pt}}
		\begin{tabular*}{\temptablewidth}{@{\extracolsep{\fill}}lcccccccc}
			\diagbox{Methods}{Datasets} & BBCS  &  3S & Washington & BBC4  & Reu & Cal7 & HW2  & CS   \\
			\hline
			C$^{2}$IMUFS   &$\mathbf{78.05\pm3.37}$  &$\mathbf{65.09\pm5.33}$   &$\mathbf{67.13\pm2.24}$       &$\mathbf{80.78\pm5.16}$   &$\mathbf{51.32\pm4.54}$ &$\mathbf{46.24\pm3.36}$  &$\mathbf{57.08\pm2.86}$   &$\mathbf{48.16\pm1.38}$\\
			AllFea   &$68.88\pm2.05 \bullet$  &$57.79\pm5.19 \bullet$   &$62.51\pm4.18 \bullet$       &$65.82\pm8.84 \bullet$   &$45.36\pm3.27 \bullet$ &$44.41\pm4.84 \bullet$  &$53.95\pm2.61 \bullet$   &$32.85\pm1.28 \bullet$\\
			LPscore   &$43.16\pm3.10 \bullet$  &$54.91\pm3.36 \bullet$   &$60.95\pm2.91 \bullet$       &$42.78\pm0.39 \bullet$   &$24.31\pm0.92 \bullet$ &$41.37\pm3.45 \bullet$  &$50.23\pm3.43 \bullet$   &$35.78\pm0.53 \bullet$\\
			EGCFS   &$51.52\pm4.28 \bullet$  &$55.74\pm5.39 \bullet$   &$58.06\pm2.33 \bullet$       &$60.55\pm6.17 \bullet$  &$29.04\pm1.55 \bullet$ &$44.25\pm3.89 \bullet$  &$29.43\pm1.36 \bullet$   &$35.05\pm0.30 \bullet$\\
			UDPFS   &$49.20\pm2.52 \bullet$  &$43.93\pm4.05 \bullet$   &$47.39\pm2.45 \bullet$       &$54.85\pm6.01 \bullet$   &$23.98\pm0.47 \bullet$ &$44.73\pm3.73 \bullet$  &$29.03\pm17.40 \bullet$  &$37.50\pm1.75 \bullet$\\
			RUSLP   &$62.90\pm5.13 \bullet$  &$57.26\pm6.29 \bullet$   &$60.57\pm2.26 \bullet$       &$69.27\pm4.88 \bullet$   &$29.14\pm2.00 \bullet$ &$44.01\pm2.76 \bullet$  &$32.36\pm1.85 \bullet$  &$22.58\pm0.51 \bullet$\\
			OMVFS   &$55.37\pm4.51 \bullet$  &$59.72\pm6.54 \bullet$   &$63.50\pm4.16 \bullet$       &$67.98\pm7.22 \bullet$   &$44.23\pm3.86 \bullet$ &$39.90\pm3.36 \bullet$  &$48.85\pm2.30 \bullet$   &$40.08\pm2.57 \bullet$\\
			ACSL   &$72.90\pm5.41 \bullet$  &$56.90\pm6.54 \bullet$   &$63.69\pm3.18 \bullet$       &$69.85\pm3.61 \bullet$  &$39.87\pm2.47 \bullet$ &$44.36\pm5.69 \bullet$  &$28.98\pm0.83 \bullet$   &$42.30\pm3.10 \bullet$\\
			CGMV-UFS   &$72.47\pm5.61 \bullet$  &$56.59\pm5.52 \bullet$   &$63.63\pm2.22 \bullet$       &$71.39\pm4.69 \bullet$   &$46.01\pm3.58 \bullet$ &$42.46\pm2.43 \bullet$  &$29.43\pm1.36 \bullet$   &$35.17\pm0.18 \bullet$\\
			NSGL   &$65.20\pm4.42 \bullet$  &$57.51\pm5.75 \bullet$   &$60.92\pm3.71 \bullet$       &$68.32\pm3.76 \bullet$   &$45.96\pm3.91 \bullet$ &$43.10\pm4.10 \bullet$  &$52.98\pm3.03 \bullet$   &$40.28\pm1.60 \bullet$\\
			CVFS   &$56.32\pm1.76 \bullet$  &$57.81\pm3.72 \bullet$   &$63.04\pm3.81 \bullet$       &$58.02\pm9.15 \bullet$  &$43.70\pm4.14 \bullet$ &$45.22\pm4.08 \circ$  &$43.63\pm1.93 \bullet$   &$40.22\pm3.47 \bullet$\\
			CvLP-DCL   &$69.68\pm3.01 \bullet$  &$59.21\pm5.95 \bullet$   &$59.56\pm1.10 \bullet$       &$70.22\pm3.90 \bullet$  &$42.21\pm4.14 \bullet$ &$45.11\pm3.29 \circ$  &$34.13\pm2.62 \bullet$   &$39.17\pm3.06 \bullet$\\
			
			\bottomrule
		\end{tabular*}
	\end{flushleft}
\end{table*}

\begin{table*}[!htbp]\huge 
	\tabcolsep 0pt
	\caption{Means and standard deviation (\%) of NMI for different methods on eight datasets with missing ratio 0.3 while selecting 20\% of all features.}\small\label{Table3}
	\vspace*{-5pt}
	\begin{flushleft}
		\def\temptablewidth{\textwidth}
		{\rule{\temptablewidth}{1pt}}
		\begin{tabular*}{\temptablewidth}{@{\extracolsep{\fill}}lcccccccc}
			\diagbox{Methods}{Datasets} & BBCS  &  3S & Washington & BBC4  & Reu  & Cal7 & HW2 & CS   \\
			\hline
			C$^{2}$IMUFS   &$\mathbf{59.99\pm3.60}$  &$\mathbf{57.68\pm4.03}$   &$28.44\pm3.24$       &$\mathbf{61.17\pm2.50}$  &$\mathbf{29.95\pm2.70}$ &$\mathbf{41.31\pm1.99}$  &$\mathbf{48.03\pm1.34}$   &$\mathbf{18.92\pm0.84}$\\
			AllFea   &$53.42\pm2.02 \bullet$  &$52.25\pm2.45 \bullet$   &$27.65\pm2.80 \circ$       &$52.31\pm8.84 \bullet$  &$25.32\pm2.63 \bullet$ &$38.18\pm2.23 \bullet$  &$46.59\pm1.11 \bullet$   &$10.60\pm0.78 \bullet$\\
			LPscore   &$15.90\pm2.79 \bullet$  &$46.55\pm2.00 \bullet$   &$26.42\pm1.52 \bullet$       &$19.44\pm0.39 \bullet$  &$5.18\pm0.56 \bullet$ &$35.80\pm0.76 \bullet$  &$39.76\pm1.43 \bullet$   &$17.04\pm0.23 \bullet$\\
			EGCFS   &$27.40\pm4.40 \bullet$  &$47.35\pm3.52 \bullet$   &$12.67\pm1.15 \bullet$       &$42.87\pm6.17 \bullet$  &$12.98\pm0.94 \bullet$ &$38.69\pm1.73 \bullet$  &$19.14\pm0.54 \bullet$   &$11.44\pm0.18 \bullet$\\
			UDPFS   &$27.69\pm3.31 \bullet$  &$35.81\pm2.03 \bullet$   &$10.28\pm1.12 \bullet$       &$35.28\pm6.01 \bullet$  &$3.38\pm0.39 \bullet$ &$37.66\pm1.68 \bullet$  &$17.92\pm0.88 \bullet$   &$11.63\pm1.18 \bullet$\\
			RUSLP   &$47.58\pm3.22 \bullet$  &$51.89\pm3.96 \bullet$   &$25.79\pm4.98 \bullet$       &$51.02\pm4.88 \bullet$  &$10.56\pm1.30 \bullet$ &$36.82\pm1.60 \bullet$  &$19.98\pm0.87 \bullet$  &$10.94\pm0.07 \bullet$\\
			OMVFS   &$37.53\pm3.05 \bullet$  &$51.60\pm3.68 \bullet$   &$27.28\pm1.47 \circ$       &$50.04\pm7.22 \bullet$  &$24.42\pm2.98 \bullet$ &$33.69\pm2.09 \bullet$  &$39.58\pm1.36 \bullet$   &$16.07\pm1.14 \bullet$\\
			ACSL   &$55.49\pm4.29 \bullet$  &$50.54\pm3.91 \bullet$   &$27.73\pm4.30 \circ$       &$51.05\pm3.61 \bullet$  &$22.16\pm1.48 \bullet$ &$32.01\pm2.33 \bullet$  &$17.52\pm0.48 \bullet$  &$18.85\pm1.74 \circ$\\
			CGMV-UFS   &$55.03\pm4.26 \bullet$  &$49.94\pm3.39 \bullet$   &$22.77\pm4.30 \bullet$       &$53.17\pm3.61 \bullet$  &$26.78\pm1.48 \bullet$ &$34.99\pm2.33 \bullet$  &$19.14\pm0.48 \bullet$   &$11.56\pm1.74 \bullet$\\
			NSGL   &$50.99\pm2.83 \bullet$  &$50.24\pm2.85 \bullet$   &$27.00\pm1.97 \bullet$       &$51.35\pm3.76 \bullet$  &$26.50\pm2.59 \bullet$ &$36.76\pm1.71 \bullet$  &$40.94\pm1.58 \bullet$   &$14.38\pm1.01 \bullet$\\
			CVFS   &$33.20\pm1.24 \bullet$  &$49.50\pm3.04 \bullet$   &$\mathbf{30.29\pm1.36} \circ$       &$39.92\pm9.15 \bullet$  &$24.65\pm2.53 \bullet$ &$35.47\pm1.52 \bullet$  &$34.18\pm1.03 \bullet$   &$14.04\pm1.26 \bullet$\\
			CvLP-DCL   &$55.00\pm2.36 \bullet$  &$53.25\pm4.45 \bullet$   &$22.97\pm1.32 \bullet$       &$51.49\pm3.90 \bullet$  &$23.85\pm2.66 \bullet$ &$38.23\pm1.92 \bullet$  &$23.10\pm1.14 \bullet$  &$14.19\pm1.35 \bullet$\\

			\bottomrule
		\end{tabular*}
	\end{flushleft}
\end{table*}

\subsection{Parameter Sensitivity}
\begin{figure*}[!htbp]
\centering
\includegraphics[width=\textwidth]{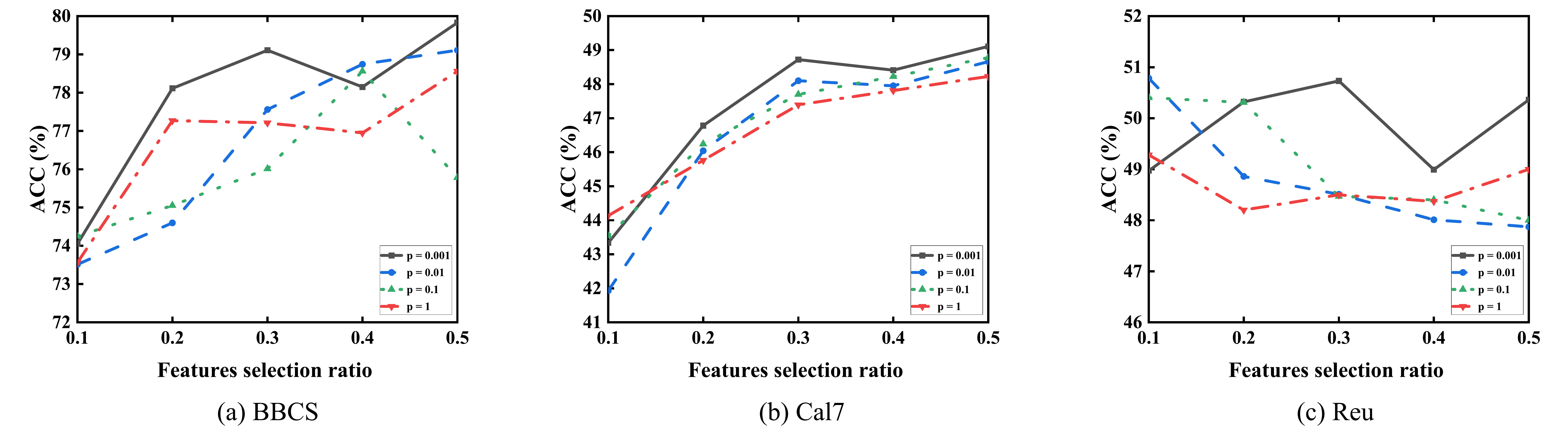}
\caption{ACC of different $p$ values with different feature selection ratios  on BBCS, Cal7 and Reu datasets.}\label{sensitivity-p}
\end{figure*}

\begin{figure*}[!htbp]
\centering
\includegraphics[width=\textwidth]{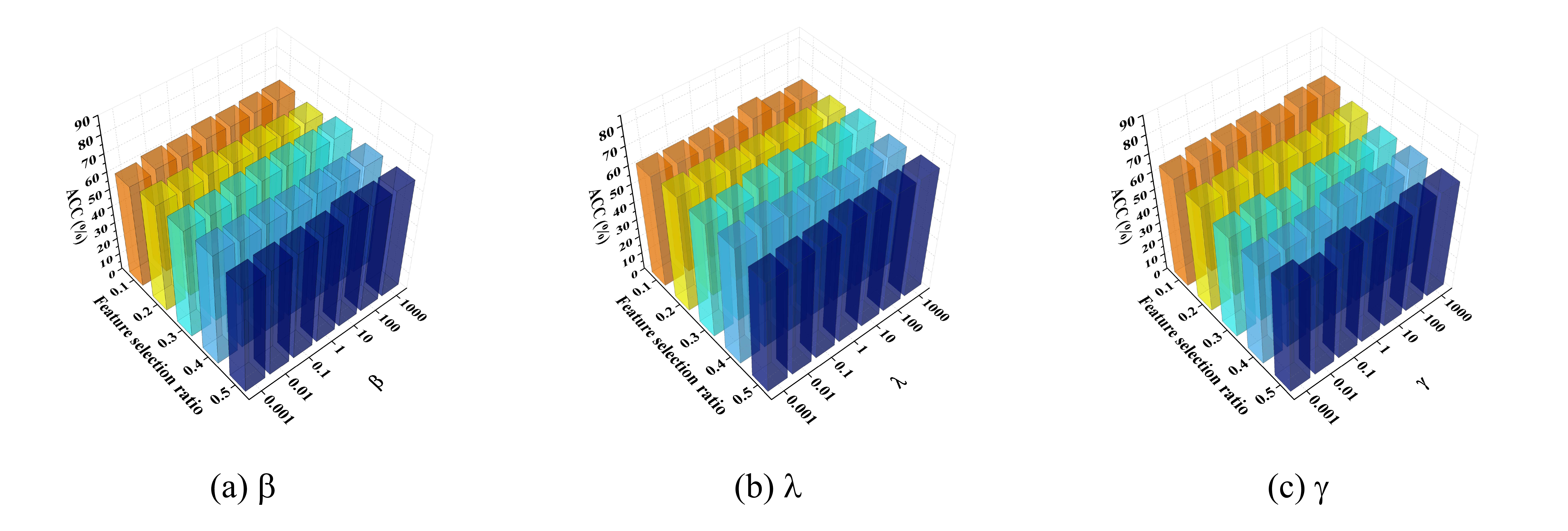}
\caption{ACC with varying parameters $\beta$, $\lambda$, $\gamma$ and feature selection ratios on BBCS dataset.}\label{sensitivity-acc}
\end{figure*}

In this subsection, we investigate the parameter sensitivity of the proposed method C$^{2}$IMUFS with regards to $p$,  $\beta$, $\lambda$ and $\gamma$. $p~(0<p\leq 1)$ is the parameter to control the sparsity and convexity of C$^{2}$IMUFS, namely, the smaller $p$ is, the  more sparse solution is. We first study the effect of $p$ value on our model performance. Fig.~\ref{sensitivity-p} shows  ACC on  BBCS, Cal7 and Reu datasets when  $p$ varies in $\{0.001,0.01,0.1,1\}$ and other parameters are tuned to obtain the optimal result.  There yields the similar results on the other datasets. From Fig.~\ref{sensitivity-p}, we can see that the performance of C$^{2}$IMUFS with $p<1$ is superior that of C$^{2}$IMUFS with $p=1$ in most of cases. Hence, we can flexibly select the value of $p$ on different datasets to obtain the more accurate sparse, which is beneficial to  improve the performance of C$^{2}$IMUFS. In addition, to investigate the sensitivity of C$^{2}$IMUFS w.r.t. parameters $\beta$, $\lambda$ and $\gamma$, we show the performance of C$^{2}$IMUFS by varying a parameter and fixing the rest two parameters. Due to the limited space, we only report the ACC of C$^{2}$IMUFS on BBCS dataset, as shown in Fig.~\ref{sensitivity-acc}. We can see that ACC show a slight fluctuations w.r.t. $\beta$ and $\lambda$ in subfigures (a) and (b), and a relatively large fluctuations w.r.t. $\gamma$ in subfigure (c). Thus, C$^{2}$IMUFS is not very sensitive to $\beta$ and $\lambda$ and relatively sensitive to $\gamma$. We need tune the parameter  $\gamma$ to obtain a better performance by using grid search method in practice.

\subsection{Convergence Analysis}
\begin{figure*}[!htbp]
\centering
\includegraphics[width=\textwidth]{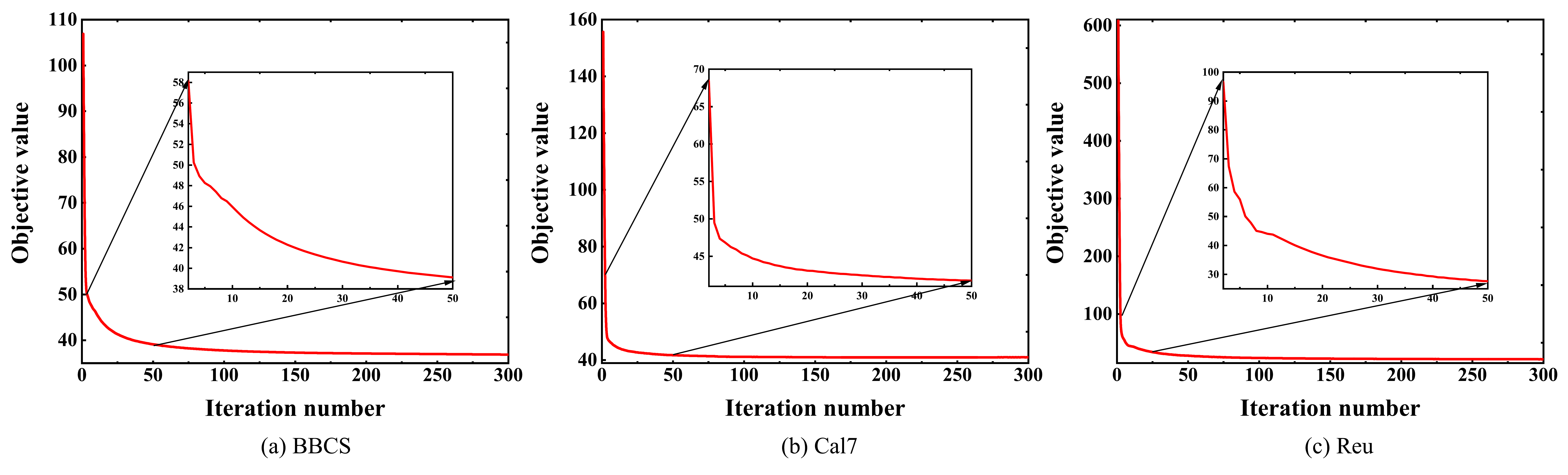}
\caption{Convergence curves of C$^{2}$IMUFS on 3S, Cal7 and Reu datasets.}\label{convergence}
\end{figure*}

We have theoretically proven the convergence of the proposed algorithm C$^{2}$IMUFS  in Section~\ref{sec:convergence}. In this subsection, we discuss its convergence speed experimentally. Fig.~\ref{convergence} shows the convergence curves of C$^{2}$IMUFS on  BBCS, Cal7 and Reu datasets, where the $x$-coordinate pertains to the number of iterations and the $y$-coordinate pertains to the objective function values. We can observe that the convergence curve decreases sharply within a few iterations and achieves stable almost within  100 iterations.

\section{Conclusions}\label{sec:Conclusions}
In this paper, we proposed a novel multi-view unsupervised feature selection method C$^{2}$IMUFS  for incomplete multi-view data.  Different from the existing methods limited the complete-view scenario, our method integrates the feature selection of incomplete multi-view data and the learning of complementary and consensus  information across different views into a unified framework. An extended WNMF model equipped with the adaptive view-weight learning and a flexible sparse constraint  was presented to select features from the incomplete multi-view data. Meanwhile, the complementary and consensus learning-guided module was developed to obtain the complete similarity graph in each view and the common clustering  indicator matrix  across of different views. We also developed an iterative optimization algorithm to solve the proposed problem with proven convergence. Experimental results on eight real-world multi-view datasets demonstrated the superiority of the proposed method compared with the state-of-the-art methods.

\section*{Acknowledgments}
This work is supported by the  Youth Fund Project of Humanities and Social Science Research of Ministry of Education (No. 21YJCZH045), National Science Foundation of China (Nos. 62176221, 62076171), the Natural Science Foundation of Fujian Province (No. 2020J01800) and the Joint Lab of Data Science and Business Intelligence at Southwestern University of Finance and Economics.

\end{document}